\pdfoutput=1

\documentclass[11pt]{article}

\usepackage[final]{acl} 

\usepackage{times}
\usepackage{latexsym}

\usepackage{tikz}
\usepackage{amsmath}
\usepackage{amssymb}
\usepackage{amsfonts}
\usepackage{subcaption}
\usepackage{multirow}
\usepackage{booktabs}
\usepackage{arydshln}
\usepackage{amssymb}
\usepackage{tikz}
\usepackage{booktabs}
\usepackage{hyperref}
\usepackage{cleveref}
\usepackage{adjustbox}
\usepackage{worldflags}
\usepackage{colortbl}
\usepackage{soul}
\usepackage{color}

\DeclareRobustCommand{\hlcyan}[1]{{\sethlcolor{gray!25}\hl{#1}}}
\newcommand\blfootnote[1]{
    \begingroup
    \renewcommand\thefootnote{}\footnote{#1}
    \addtocounter{footnote}{-1}
    \endgroup
}
\newcolumntype{g}{>{\columncolor{gray!25}}c}

\usepackage[T1]{fontenc}

\usepackage[utf8]{inputenc}

\usepackage{microtype}

\usepackage{inconsolata}

\usepackage{graphicx}

\title{\textit{Retrieve}, \textit{Generate}, \textit{Evaluate}:  A Case Study for Medical Paraphrases Generation with Small Language Models }

 \author{Ioana Buhnila*\textsuperscript{$\spadesuit$}, Aman Sinha*\textsuperscript{$\spadesuit, \clubsuit$} \and Mathieu Constant\textsuperscript{$\spadesuit$} \\
         \textsuperscript{$\spadesuit$}ATILF, CNRS, Université de Lorraine, Nancy, France \\
         \textsuperscript{$\clubsuit$}Institut de Cancérologie, Strasbourg, France \\
           \texttt{firstname.lastname@univ-lorraine.fr} \\}

\begin{document}
\maketitle
\begin{abstract}

Recent surge in the accessibility of large language models (LLMs) to the general population can lead to untrackable use of such models for medical-related recommendations. Language generation via LLMs models has two key problems: firstly, they are prone to hallucination and therefore, for any medical purpose they require scientific and factual grounding; secondly, LLMs pose tremendous challenge to computational resources due to their gigantic model size. In this work, we introduce \textbf{pRAGe}, a \textbf{p}ipeline for \textbf{R}etrieval \textbf{A}ugmented \textbf{G}eneration and \textbf{e}valuation of medical paraphrases generation using Small Language Models (SLM). We study the effectiveness of SLMs and the impact of external knowledge base for medical paraphrase generation in French.


\end{abstract}

\blfootnote{*These authors contributed equally to this work.}

\section{Introduction}

Large Language Models (LLMs) are used for a big variety of NLP tasks and are found to be very effective, but they exhibit a specific trait that can make them unreliable: \textit{hallucinations} (\citealt{zhang2023siren}; \citealt{huang2023survey}). LLM hallucinations are incorrect output generations that do not correspond to the input prompt, or are not factual information reflecting world knowledge. LLMs such as ChatGPT \cite{OpenAI2022} became widely used by lay people and the risk of incorrect information spreading within medical text generation persists. In very specialized and sensitive fields such as medicine, hallucinations can have dangerous consequences for the patient via wrong prognosis or treatment recommended by LLMs \cite{umapathi2023med}. This leads to the necessity that the information patients receive to be scientifically and factually grounded, either by human expert or external knowledge base. 
Additionally,  recent state-of-the-art (SOTA) results are using LLMs with dozens or hundreds of billion parameters. Their results cannot be easily reproduced due to high costs of GPUs for finetuning and inference. GPT-4 \cite{achiam2023gpt} is at the top for many NLG tasks, but the model is not open-source and API usage can become costly during experiments.

\begin{table}
\centering
    \resizebox{\columnwidth}{!}{
    \begin{tabular}{c|p{2cm}p{5cm}}
\hline
\textbf{} & \textbf{Term} & \textbf{Paraphase}\\
\hline
\textbf{fr} & {hypopnée} & {respiration partiellement bloquée} \\
\textbf{en} & {hypopnea} & {partially blocked breathing} \\
\hline
\textbf{fr} & {myasthénie grave} & {est un trouble qui entraîne une faiblesse musculaire et une fatigue musculaire excessive} \\
\textbf{en} & {myasthenia gravis} & {is a condition that leads to muscle weakness and excessive muscle fatigue} \\
\hline
\textbf{fr} & {akathisie} & {agitation intérieure et incapacité à rester assis} \\
\textbf{en} & {akathisia} & {inner restlessness and inability to sit still} \\
\hline
\end{tabular}}
\caption{Examples of medical term paraphrase in French (fr) and its translation in English (en) from RefoMed dataset. Each term is a medical term automatically identified with SNOMED-3.5VF, and corresponding paraphrase represents a sub-sentential paraphrase.}
\label{tab:refomed-examples}
\end{table}
Several works such as \cite{leblanc2014patient, tavakoly2020communication} report frequent misunderstanding which caused due to medical jargon (i.e. medical terms) usage which are highly present in doctor-patient interactions because patients' different health-literacy levels. Such terms are lexical units that designate a concept from a specialised domain \cite{condamines97}. These complex terms have to be adapted or simplified for lay people through paraphrases, short definitions or explanations. We consider that short sequences of words (shorter than a sentence) can provide fast explanations adapted to patients needs of understanding as shown in Table \ref{tab:refomed-examples}.   

Therefore, in this work we aim to develop a method of medical term paraphrase generation that can help patients and their families better understand and follow the treatment of their illness. In particular, we focus on the generation of \textbf{sub-sentential paraphrases}, which are defined as simple words or sequences of words \cite {bouamor2013multitechnique, max2012generalizing} for better understanding of technical terms.  
We introduce \textbf{pRAGe}, an augmented SLM (Small Language Model) pipeline that generates paraphrases and short definitions for medical terms based on patient query. We make use of RALMs (Retrieval Augmented Language Models), which combine a RAG architecture and a language model (LM). RAG (Retrieval Augmented Generation) \cite{lewis2020retrieval} models help reduce the level of hallucinations generated by LMs by accessing an external knowledge base (KB) to retrieve the answer for an input prompt. 

We conduct our RAG experiments on a downstream natural language generation (NLG) task: medical paraphrase generation \cite{gupta2018deep}, paired with a question-answering (Q\&A) task \cite{singhal2023large}. Our paper focuses on open-source Small Language Models \cite{schick2021s} and cost efficient 
quantization methods that can allow our experiments to be easily reproduced by the community. Hence, we aim to answer the following research questions: 
\begin{itemize}
\item \textbf{RQ1}: \textit{How good are open-source small language models (1 to 7B parameters) quantized VS fine-tuned models at medical Q\&A task in a RAG system? }
\item \textbf{RQ2}: \textit{What is the impact of finetuning VS prompting in the medical paraphrase generation task? }
\item \textbf{RQ3}: \textit{How can we evaluate the quality of RAG systems for medical paraphrase generation?  }

\end{itemize}

\begin{figure*}[t]
    \centering
    \includegraphics[trim = 0cm 0.cm 3cm 0cm, scale=0.8]{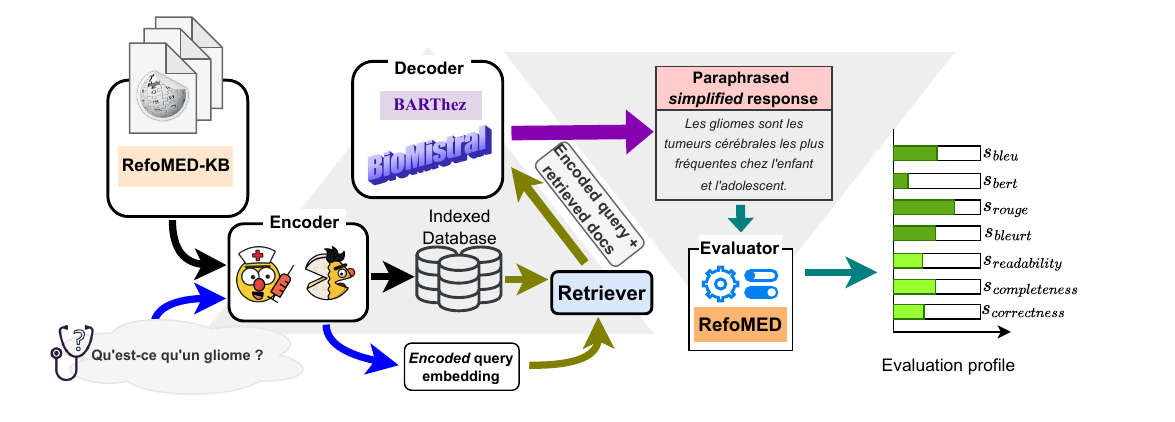}
    
    \caption{Illustration of pRAGe experimental pipeline. The illustration is intended to read from left to right. Each colored arrow represent a process.  The \textcolor{black}{$\blacksquare$} arrow indicates the \textbf{creation of indexed database}; the \textcolor{blue}{$\blacksquare$} arrow indicates the \textbf{encoding of the query}; the \textcolor{olive}{$\blacksquare$} arrow represents \textbf{retrieval of relevant documents}; the \textcolor{violet}{$\blacksquare$} arrow denotes the \textbf{generation of simplified paraphrase output} and the \textcolor{teal}{$\blacksquare$} arrow indicates the \textbf{evaluation step for the generated output} to obtain the evaluation profile of the paraphrase. }
    \label{fig:input_output}
\end{figure*}

The main contributions of our paper are:  
\textbf{(1)} \textbf{pRAGe} (\textit{\textbf{p}ipeline for \textbf{R}etrieval \textbf{A}ugmented \textbf{G}eneration and \textbf{e}valuation}), an open source RAG pipeline for medical paraphrase and explanation generation in a Q\&A downstream task making use of only non proprietary LLMs;  
\textbf{(2)} \textbf{RefoMed-KB} (\textit{Medical Paraphrases Knowledge Base)}, an French medical knowledge base about scientific medical terms extracted from the RefoMed dataset\footnote{https://github.com/ibuhnila/refomed};  
\textbf{(3)} \textbf{pRAGe-FT}, a fine-tuned RAG model\footnote{https://huggingface.co/amasi/biomistral-gptq-ft} for the paraphrase generation task in French and an evaluation of its performance on zero-shot inference Q\&A task. We share our code, datasets and evaluation metrics with the NLP community to support open-access and reproducible research\footnote{https://github.com/ATILF-UMR7118/pRAGe/}.

\section{Related Work}

SOTA Large Language Models such as GPT-3.5 \cite{OpenAI2022}, GPT-4 \cite{achiam2023gpt}, Mistral and Mixtral \cite{jiang2023mistral} and Llama-2 \cite{touvron2023llama} can give impressive results on question-answering tasks \cite{tan2023can}, or radiology reports simplification \cite{jeblick2023chatgpt}. However, these models are either not open-source (GPT) or may require access to powerful servers. Smaller models (<7B) are easier and feasible to implement in downstream tasks, but they tend to hallucinate more compared to LLMs. In health related applications, we need to retrieve correct information, thus RAG systems are essential.  

\paragraph{Different types of RAG systems} have been developed \cite{gao2023retrieval}, going from the original naive RAG (simple structure of a retriever and a generator) \cite{lewis2020retrieval} to more advanced or modular RAG such as RA-DIT \cite{lin2023ra}. There are several types of advanced RAG architectures such as the self-reflective Self-RAG \cite{asai2023self}, Self-BioRAG \cite{jeong2024improving}, black box RAG like RePlug  \cite{shi2023replug}, pretrained model RAG such as REALM \cite{guu2020retrieval}, RETRO \cite{borgeaud2022improving} or ATLAS \cite{izacard2022atlas}. However, these advanced RAG systems have very complex implementation architectures and require servers with many GPUs.
Almanac \cite{zakka2024almanac} is a RAG system developed for the clinical domain  , but it is not open source and it is available only for English. We are interested in developing easy to implement RAG systems for languages other than English, in our case for French. There are some French language models, such as CamemBERT \cite{martin2019camembert} for the general domain and DrBERT \cite{labrak2023drbert} for the medical domain, but they are not adapted for text-to-text generation tasks.

\paragraph{Multilingual decoder models,} such as Mixtral, Falcon, Bloom, and French models such as Vigogna (based on Vicuna) or Claire  are available for general language. As for encoder-decoder French models trained for NLG task (more precisely for summarization tasks), we cite BARThez \cite{eddine2020barthez}, fine-tuned on BART. Many open-source medical LLMs for NLG tasks are available for English, such as PMC-Llama \cite{wu2024pmc}, MediTron \cite{chen2023meditron} (both based on LLama-2), ClinicalGPT \cite{wang2023clinicalgpt}, but there is no NLG medical model trained exclusively on the French language. Multilingual medical models such as Medical-mT5 \cite{garcia2024medical} and BioMistral \cite{labrak2024biomistral} are of interest for our research.  
One important aspect when working with LLMs on medical text is the quality of the generation in order to minimize the hallucination risk. LLMs can generate content according to a specific prompt.  

\textbf{However, we want to improve the quality} of the output and obtain a short paraphrase of the term and not a complete description of the term (as shown in Table \ref{tab:refomed-examples}). Thus, we use prompt tuning 
techniques. However, we need curated sub-sentential paraphrase datasets in medical French for this task. Most of the paraphrase datasets contain only sentential paraphrases from general language in English: MSRP
\cite{dolan2004unsupervised}, PPDB \cite{ganitkevitch2014multilingual}, PAWS \cite{zhang2019paws} or multilingual: TaPaCo \cite{scherrer2020tapaco} or ParaCotta \cite{aji2022paracotta}.  

\textbf{French sentential paraphrase databases} are scarce. One such resource for the medical domain is the WikiLarge FR and CLEAR, a parallel corpus for text simplification built with translation techniques \cite{cardon2020french}. However, there is little work on sub-sentential paraphrases, as they are difficult to identify. Previous work explored crowdsourcing methods \cite{tschirsich2013leveraging} or translation techniques \cite{bouamor2013multitechnique} \cite{zhai2020building}. A dataset similar to our work is PARADE \cite{he2020parade}, containing computer science definition-style paraphrases for English technical concepts extracted from online user-generated flashcards. 
 

\section{Methodology}
\begin{table}[!t]
    \centering
    \resizebox{\columnwidth}{!}{
    \begin{tabular}{c|cc}
        \hline
        \hline
        \textbf{SLM} &\multicolumn{2}{c}{\textbf{FR Encoder-Decoder }}  \\
        \hline
        \hline
         \textbf{BaseSLM} &\multicolumn{2}{c}{\texttt{BARThez-orangesum-abstract}}\\ 
         \textbf{} &\multicolumn{2}{c}{\texttt{BioMistral-7B-SLERP-GPTQ}}\\ 
         \hline
         \hline
        \textbf{pRAGe} &\textbf{FR Encoder} &\textbf{FR Decoder} \\
        \hline
        \hline
         \textbf{BioMistral}& \texttt{DrBERT}&\texttt{BioMistral-7B-SLERP-GPTQ}\\
        \textbf{pRAGe} & \texttt{sent-CamemBERT}&\texttt{BioMistral-7B-SLERP-GPTQ}\\
         \hline
         \textbf{BARThez} & \texttt{DrBERT}&\texttt{BARThez-orangesum-abstract}\\
         \textbf{pRAGe}  & \texttt{sent-CamemBERT}&\texttt{BARThez-orangesum-abstract}\\
        \hline
    \end{tabular}}
    \caption{Configurations of non proprietary French encoders and decoders tested in our experiments.}
\end{table}
We illustrate our method in Figure 1. 
\textbf{pRAGe} is built on a encoder-retriever-decoder framework. We designed \textbf{pRAGe} to embed medical query and to generate an output in a style that \textit{translates} medical knowledge for patients in a simpler language, e.g. \textit{rhizarthrose $\rightarrow$ arthrose du pouce} (rhizarthrosis $\rightarrow$ arthrosis of the thumb). Therefore, we pair models for the general language with medical models. We tested different configurations of non proprietary encoders and decoders, as shown in Table 2. We used the general French encoder model \textbf{sent-CamemBERT} (\citealt{reimers2019sentence, martin2020camembert}) and the domain specific model \textbf{DrBERT} \cite{labrak2023drbert}, a French BERT type model for the medical field. The \textbf{pRAGe} pipeline encodes in embeddings the input query and the Wikipedia knowledge base, RefoMed-KB (to be presented in section 3.1).  
We did prompt engineering to guide the decoder towards the expected output in both experimental settings, base SLM inference and pRAGe pipeline. The task attributed to the SLM is "to answer the user's question with a paraphrase, explanation or short definition" (full prompts in Figure 2 and in Appendix A.1). We used \textbf{BARThez-OrangeSum-abstract} (BARThez\footnote{For readability reasons, we will hence refer to BARThez-OrangeSum-abstract as BARTHEZ.}) \cite{eddine2020barthez}, a French seq2seq SLM, and \textbf{BioMistral-7B-SLERP-GPTQ}\footnote{BioMistral was pre-trained on 3 bilion tokens data from PubMed Central from Mistral. Less than 1,25\% of the data is a GPT-3.5 Turbo automatic translation in French and other 8 languages.}  \cite{labrak2024biomistral}, a 4-bit precision GPTQ quantized \cite{frantar2022gptq} multilingual medical model for training and inference efficiency\footnote{https://huggingface.co/LoneStriker/BioMistral-7B-SLERP-GPTQ}. We chose the BioMistral-7B-SLERP model \cite{shoemake1985animating} as it gave the best benchmark results on French datasets, according to the authors of the model \cite{labrak2024biomistral}. 
We also tested the impact of \textbf{finetuning the SLMs} in our \textbf{pRAGe} system on an existing sub-sentential paraphrase dataset in medical French, \textbf{RefoMed} \cite{buhnila2023methode}. For finetuning we used the \textbf{Q-LoRA} method \cite{dettmers2024qlora}, a computational efficient finetuning method that reduces the number of parameters for BioMistral from 7B to 1,38B parameters. 
We present the RefoMed dataset in the next section and the evaluation metrics for \textbf{pRAGe} generated paraphrases in section 3.2.

\subsection{Datasets}

We split the RefoMed dataset for finetuning in training, validation and test sets. The validation and test sets were used to built the knowledge base for the RAG system, RefoMed-KB. We present both datasets below.

\paragraph{RefoMED}
We used an unique open-source dataset of medical sub-sentential paraphrases in French, \textbf{RefoMed}\footnote{https://github.com/ibuhnila/refomed} \cite{buhnila2023methode} as input queries and to finetune the SLMs in \textbf{pRAGe}. The RefoMed corpus is made of 6,297 pairs of unique medical terms and their corresponding sub-sentential paraphrases. The source corpora are \textbf{ClassYN} \cite{todirascu2012french} and \textbf{CLEAR Cochrane} \cite{grabar2018clear}, both comparable corpora of scientific and simplified medical short texts in French.  
The RefoMed dataset was built by automatically extracting sentences that contain medical terms and paraphrases from the source corpora. The author \cite{buhnila2023methode} identified medical terms automatically by using a rule-based method and the \textbf{SNOMED-3.5VF} French medical terminology \cite{cote1998systematized}.The paraphrases were identified with the help of linguistic paraphrase markers such as \textit{c'est-à-dire} ("so called"), \textit{également appelé} ("also called"), \textit{est une maladie} ("is a disease"), and punctuation signs, such as colons and brackets (\citealt{grabar2015extraction}; \citealt{antoine2016exploitation}; \citealt{buhnila2022role}).  

\paragraph{In order to avoid bias in the LM's finetuning,} we split the dataset by unique term entry while staying in the range of the classic 60-20-20 train-validation-test split proportion. This split was important because in the RefoMed dataset we can find multiple paraphrases for one particular term. For instance, the term "placebo" has various paraphrases: (1) "absence d'intervention" / no intervention; (2) "médicament inactif" / inactive drug; (3) "traitement factice" / fake treatment; (4) "par exemple une pilule de sucre" / for example a sugar pill; (5) "aucun traitement" / no treatment.  
Thus, the resulting split is as follows: 3,981 term-paraphrase pairs for training, 1,063 for validation and 1,253 pairs for testing. 

\paragraph{Descriptive statistics for the paraphrases in the RefoMed dataset.} We counted the length of the paraphrases in RefoMed. The shortest paraphrase is of 1 word length whereas the longest is 83 words length. The mean and standard deviation is 10.34 and 8.15 respectively. Accounting the above, we consider 10, 25 and 50 word count as the limit of paraphrase generation for our various RAG systems. For our final analysis, we considered token limit of 25 and 50.

\paragraph{RefoMED-KB}The next step was to build the knowledge base (KB) for the medical terms from the validation and testing sets. We automatically extracted top-3 Wikipedia articles where the terms appear in the title of the article using the Python \textit{wikipedia} library. We extracted the first 20 lines of each relevant wikipedia page and we obtain a medical knowledge base in French of 20,402 sentences (1,708,034 tokens) about the 1,253 medical terms from the test list. 

\subsection{Automatic Evaluation}

We develop an evaluation method for your system to tackle \textbf{(RQ3)}: \textit{How can we evaluate the quality of RAG systems medical paraphrase generations?}
Evaluation for the complete RAG framework can be divided into two categories: intrinsic and extrinsic. For extrinsic evaluation, we check for hallucination by evaluating the rate of medical correct answers \cite{huang2023survey}. For intrinsic evaluation, we perform manual evaluation by checking the quality of responses generated.

\paragraph{\textbf{Evaluation metrics.}} Several metrics are used in SOTA research on text generation evaluation: \textbf{ROUGE} \cite{lin2004rouge}, calculates the n-grams overlap (recall), \textbf{BLEU} \cite{papineni2002bleu}, computes the number of similar n-grams between the output and the reference (precision), \textbf{BERTscore} \cite{zhang2019bertscore}, compares the embeddings of tokens that match in the output and reference text, while \textbf{BLEURT} \cite{sellam2020bleurt}, computes the semantic similarity and lexical difference between them. \textbf{MEDCON} \cite{yim2023aci} is a metric that computes the F1-score of the UMLS concepts found both in the output and the reference text (however, available only for English). 
In our work, we want to evaluate the similarity between the generated output text and the reference text in French. We use the following evaluation metrics: \texttt{bleu}, \texttt{rouge}, \texttt{bleurt}, and \texttt{bertscore}. 


\paragraph{RAG\(ref\)S ($S$)} We define a score metric for evaluating the generated response set from pRAGe. For any $\text{i}_{th}$ query, if the $p_i$ is the generated response from the RAG pipeline and $\mathcal{R}_i$ is the list of reference paraphrases. 

\begin{equation}
    S_{\Omega} = \frac{\sum_{i=1}^N \texttt{max}(\{\Omega(p_i, r_{ij}) \forall  r_{ij} \in \mathcal{R}_i \}) }{N} 
    \label{eqn:rag-eval}
\end{equation}

where $N$ is number of queries and $\Omega$ is a lexical or semantic similarity comparison metric such as \texttt{bleu}, \texttt{rouge}, \texttt{bleurt}, \texttt{bertscore}, etc. 

\subsection{Fine-grained Human Evaluation}
We conduct a fine-grained evaluation of the generated paraphrases. Firstly, we automatically evaluate the generation quality with SOTA metrics (\texttt{bleu}, \texttt{rouge}, \texttt{bleurt}, \texttt{bertscore}) with our metric, RAG\(ref\)S ($S$) as introduced in section 3.2. In addition to the overall generation quality, we study the individual generation quality to obtain more insights. A set of 1200 examples\footnote{50 examples from 24 different configurations, as presented in Table 7.} were manually analyzed by 3 French proficient linguist annotators following different criteria: 
    \paragraph{- \texttt{\textbf{readability:}}} scored from \textit{1} to \textit{3}, where \textit{1} means that the generated text is fluent, grammatically correct and easy to understand for laypeople; \textit{2} means the generated text includes invented words, English words or grammatical mistakes, or scientific terms in a grammatically correct context; \textit{3} represents generated text that has the incorrect traits of score \textit{2}, plus scientific terms, rendering the text difficult to understand for laypeople;
    \paragraph{- \texttt{\textbf{completeness:}}} the generated text represents a full answer, meaning the language model generated a concise answer (score \textit{1} if the text respects this condition, \textit{0} if not). We annotated two types of \texttt{completeness: \textbf{relaxed}} - the generated text contains one incomplete sentence or it contains a second incomplete sentence, and \texttt{\textbf{strict}} - the generated text contains one syntactically independent sentence;
    \paragraph{- \texttt{\textbf{correctness:}}} the generated text encompasses the correct medical knowledge and it is in French\footnote{During our validation experiments we noticed the presence of English answers in the generated answers, especially for BioMistral (the model has less then 1\% of its training data in French.)} (score \textit{1} if the two conditions are fulfilled, \textit{0} if not). We considered two types of \texttt{correctness}: \texttt{\textbf{relaxed}} - the general meaning of the medical term is comprehensible from the generated text, and \texttt{\textbf{strict}} - the exact meaning is both comprehensible and complete.

\section{Experiments}

In this section, we describe the experiments we conducted to study the comparison between SLMs and pRAGe models for generating paraphrases for medical terms. We also describe the fine-grained evaluation process in the subsections below.

\subsection{SLMs Zero-Shot Inference \textit{VS} pRAGe}

\begin{table*}[!t]
    \centering
    \resizebox{\textwidth}{!}{%
    \begin{tabular}{crl| cccc c cccc}
        \toprule
        &\multicolumn{2}{c|}{\multirow{2}{*}{\textbf{Model Setup}}}& \multicolumn{4}{c}{\textbf{Tokens=25}} && \multicolumn{4}{c}{\textbf{Tokens=50}}\\
        \cline{4-7}
        \cline{9-12}
        &&& \texttt{bert} &\texttt{bleurt} & \texttt{bleu-1}&\texttt{rouge-1}  && \texttt{bert} &\texttt{bleurt} & \texttt{bleu-1}&\texttt{rouge-1} \\
        \midrule
        \rowcolor{gray!25}
    &&&
    \multicolumn{9}{c}{w/o FINE TUNING} \\
    \parbox[t]{2mm}{\multirow{2}{*}{\rotatebox[origin=c]{90}{{\textbf{SLM}}}}} 
    &\multicolumn{2}{c|}{\texttt{BARTHEZ}} &$0.63_{{0.03}}$&$0.10_{{0.10}}$&$0.04_{{0.06}}$&$0.07_{{0.08}}$&&$0.63_{{0.03}}$&$0.10_{{0.10}}$&$0.04_{{0.06}}$&$0.07_{{0.08}}$\\
    
    & \multicolumn{2}{c|}{\texttt{BIOMISTRAL}} & $\textbf{0.70}_{{0.06}}$&$\textbf{0.15}_{{0.15}}$&$\textbf{0.11}_{{0.12}}$&$\textbf{0.20}_{{0.16}}$&&$\textbf{0.68}_{{0.06}}$&$\textbf{0.16}_{{0.15}}$&$\textbf{0.08}_{{0.08}}$&$\textbf{0.18}_{{0.13}}$\\
   
    \midrule

    \parbox[t]{2mm}{\multirow{4}{*}{\rotatebox[origin=c]{90}{{\textbf{pRAGe}}}}}& \texttt{camemBERT}&\texttt{BARTHEZ} & $0.65_{{0.05}}$&$0.07_{{0.09}}$&$0.05_{{0.07}}$&$0.12_{{0.10}}$&&$0.65_{{0.05}}$&$0.11_{{0.11}}$&$0.05_{{0.06}}$&$0.12_{{0.10}}$
\\
    
    &\texttt{DrBERT}&\texttt{BARTHEZ}&$0.64_{{0.03}}$&$0.02_{{0.06}}$&$0.04_{{0.06}}$&$0.10_{{0.09}}$ && $0.65_{{0.04}}$&$0.05_{{0.07}}$&$0.05_{{0.06}}$&$0.11_{{0.09}}$\\

    
    & \texttt{camemBERT}&\texttt{BIOMISTRAL} &$\textbf{0.69}_{{0.06}}$&$\textbf{0.14}_{{0.15}}$&$\textbf{0.12}_{{0.14}}$&$\textbf{0.19}_{{0.17}}$&&$\textbf{0.68}_{{0.06}}$&$\textbf{0.17}_{{0.15}}$&$\textbf{0.08}_{{0.09}}$&$\textbf{0.18}_{{0.14}}$
\\
    
    & \texttt{DrBERT}&\texttt{BIOMISTRAL}&
    $\textbf{0.69}_{{0.06}}$&$\textbf{0.14}_{{0.15}}$&$0.11_{{0.12}}$&$0.18_{{0.17}}$&&$\textbf{0.68}_{{0.06}}$&$\textbf{0.17}_{{0.15}}$&$\textbf{0.08}_{{0.08}}$&$0.17_{{0.13}}$\\
    
    \midrule
    \rowcolor{gray!25}
    &&&
    \multicolumn{9}{c}{w/  FINE TUNING} \\
    \midrule
    \parbox[t]{2mm}{\multirow{2}{*}{\rotatebox[origin=c]{90}{\small{\textbf{SLM}$\bigstar$}}}} 
    &\multicolumn{2}{c|}{\texttt{BARTHEZ}}&$0.62_{{0.02}}$&$0.05_{{0.08}}$&$0.06_{{0.07}}$&$0.11_{{0.08}}$ & &$0.63_{{0.03}}$&$0.09_{{0.09}}$&$0.07_{{0.07}}$&$0.12_{{0.08}}$\\
    & \multicolumn{2}{c|}{\texttt{BIOMISTRAL}} & $\textbf{0.72}_{{0.07}}$&$\textbf{0.15}_{{0.17}}$&$\textbf{0.14}_{{0.13}}$&$\textbf{0.22}_{{0.17}}$&&$\textbf{0.69}_{{0.07}}$&$\textbf{0.16}_{{0.16}}$&$\textbf{0.10}_{{0.10}}$&$\textbf{0.18}_{{0.13}}$\\
    
    \midrule
    \parbox[t]{2mm}{\multirow{4}{*}{\rotatebox[origin=c]{90}{{\textbf{pRAGe}$\bigstar$}}}}& \texttt{camemBERT}&\texttt{BARTHEZ}&$\textbf{0.65}_{{0.05}}$&$0.05_{{0.09}}$&$\textbf{0.06}_{{0.07}}$&$0.12_{{0.10}}$&
    &$\textbf{0.64}_{{0.05}}$
    &$0.10_{{0.10}}$
    &$\textbf{0.06}_{{0.07}}$
    &$\textbf{0.12}_{{0.10}}$\\
    
     &\texttt{DrBERT}&\texttt{BARTHEZ}&$0.64_{{0.03}}$&$0.01_{{0.04}}$&$\textbf{0.06}_{{0.07}}$&$\textbf{0.13}_{{0.10}}$&&$\textbf{0.64}_{{0.04}}$&$0.05_{{0.07}}$&$0.05_{{0.06}}$&$\textbf{0.12}_{{0.09}}$\\

    & \texttt{camemBERT}&\texttt{BIOMISTRAL} &$0.60_{{0.04}}$&$\textbf{0.13}_{{0.11}}$&$0.03_{{0.03}}$&$0.09_{{0.06}}$&&$0.60_{{0.05}}$&$\textbf{0.16}_{{0.11}}$&$0.03_{{0.03}}$&$0.09_{{0.06}}$\\

     & \texttt{DrBERT}&\texttt{BIOMISTRAL} &$0.59_{{0.04}}$&$0.12_{{0.15}}$&$0.03_{{0.03}}$&$0.08_{{0.06}}$&&$0.60_{{0.04}}$&$0.14_{{0.15}}$&$0.03_{{0.02}}$&$0.08_{{0.06}}$\\
     
    
    \bottomrule
\end{tabular}
    }
    \caption{Automatic Evaluation Metric Comparison of BaseSLMs with pRAGe setups on test set. Top scores for each model setups are shown in \textbf{bold}.}
    \label{tab:auto-summary}
\end{table*}

Firstly, we test SLMs ability to generate medical paraphrases in an zero-shot setting. We chose this method as it reflects real-life usage of language models by laypeople or patients. 
We test this method to answer \textbf{(RQ1)}: \textit{How good are open-source small language models and quantized models (1B-7B) at medical Q\&A alone VS in a RAG system?} We are interested in analyzing if the LM's parametric knowledge (learned during the pre-training phase) is sufficient for generating accurate paraphrases, explanations or short definitions of medical terms. We compare these results with the settings with added non-parametric knowledge, the RefoMed-KB corpus.  
For this, we test a GPTQ quantized version of BioMistral, BioMistral-7B-SLERP-GPTQ4 \cite{labrak2024biomistral}, and BARThez in an inference alone setting \cite{eddine2020barthez}. Furthermore, we test these two SLMs integrated in the pRAGe pipeline as decoders. 

\subsection{Vanilla Inference \textit{VS} Finetuning} 

RAG systems are useful in the mitigation of hallucinations, as they give extra-knowledge to the LM. However, we want to test to what extent finetuning helps the LM generate a more accurate medical paraphrases in the pRAGe pipeline (added knowledge through RefoMed-KB) and inference alone (parametric knowledge only) \textbf{(RQ2)}. We therefore test the two SLMs in two settings: non-fine-tuned (\textbf{NonFT}) and fine-tuned (\textbf{FT}) on the RefoMed paraphrase dataset.

\subsection{Implementation Details}

We tested two different lengths for the generated text: 25 and 50 tokens.
For the inference setting, we used a simple prompt in French (Figure 2) for the Base SLM and a RAG adapted prompt in French for our pRAGe pipeline (Appendix A.1). We decided to use the prompts in French, as initial test experiments with English prompts generated French-English text. 

\begin{figure}[h]
    \centering
    \begin{tikzpicture}
        \node[draw, rounded corners, minimum width=6cm, minimum height=2cm, align=left] 
        {Expliquez-moi le terme médical  \\
        en mots simples, par une paraphrase \\ 
        ou une courte définition :};
    \end{tikzpicture}
    \caption{Our prompt template in French for inference.}
    \label{fig:prompt}
\end{figure}

We present the results of our experiments and our fine-grained analysis in the results section below.

\section{Results and Discussion}

We present the summarized automatic evaluation in Table \ref{tab:auto-summary} and complete automatic evaluation table in Appendix A.2 (See Table \ref{tab:final-test-reports}). 
The automatic evaluation shows that \texttt{BIOMISTRAL} SLM reponses are more semantically and lexically related to gold paraphrase compared to \texttt{BARTHEZ} SLM. Further, both SLMs benefit from finetuning. Further, we notice that \texttt{BIOMISTRAL} pRAGe setups are obtained lower scores with finetuning. On the contrary, \texttt{BARTHEZ} pRAGe setups overall stay unaffected by fine tuning. This observation can be attributed to the fact that in pRAGe setups models are restricted by the external knowledge base whereas in the case of SLM only setup, the models are free to generate anything and therefore, can be prone to hallucinations. %

\begin{table*}[!t]
    \centering
    \resizebox{\textwidth}{!}{%
    \begin{tabular}{crl ggggg c ggggg}
        \toprule
         &&& \multicolumn{5}{c}{\textbf{w/o FINE TUNING}} &&\multicolumn{5}{c}{\textbf{w/ FINE TUNING}}\\
         \cline{4-8}
         \cline{10-14}
        \rowcolor{white}&& &   &  \multicolumn{2}{c}{$completeness$\%($\uparrow$)} &  \multicolumn{2}{c}{$correctness$\%($\uparrow$)} &&     &  \multicolumn{2}{c}{$completeness$\%($\uparrow$)} &  \multicolumn{2}{c}{$correctness$\%($\uparrow$)}\\
        \rowcolor{white}&&&\multirow{-2}{*}{$readability$ ($\downarrow$)}&\small{STRICT} & \small{RELAX}& \small{STRICT} & \small{RELAX} &&\multirow{-2}{*}{$readability$ ($\downarrow$)}&\small{STRICT} & \small{RELAX}& \small{STRICT} & \small{RELAX}\\
        \midrule

    \rowcolor{white}  &\multicolumn{2}{c}{\texttt{BARTHEZ}} &1.22& \textbf{100} & \textbf{100} & 0 & 0 & & \underline{1.36} & 0 & 0 & 0 & 0\\
    &&&  1.20 & \textbf{100} & \textbf{100}&  0 & 0 && 1.42 & 0 & 0 & 0 & 0 \\
    
    \rowcolor{white}& \multicolumn{2}{c}{\texttt{BIOMISTRAL}} &\textbf{1.08} & 10 & 20 & \underline{68} & \textbf{96} && \textbf{1.34} & \underline{16}& \underline{20} & \textbf{90} & \textbf{94}\\
    \parbox[t]{2mm}{\multirow{-4}{*}{\rotatebox[origin=c]{90}{{\textbf{SLMs}}}}}&&& 1.10 & 18 & 96 & \textbf{94} & \textbf{96} & & 1.5 & \textbf{24} & \textbf{42} & \textbf{90} & \textbf{94}\\
    \midrule

    \rowcolor{white}& \texttt{camemBERT}&\texttt{BARTHEZ} & 1.22 & 56&64& 42 &46 &&1.22 & 14 & 14&38 & 42\\
    &&& 1.26 & \textbf{96} & 96 & 46 & 50 & & 1.34 & \underline{70} & \underline{76} & 48 & 50\\
    \rowcolor{white}&\texttt{DrBERT}&\texttt{BARTHEZ} & \textbf{1} & 18 & 68 & 0 & 0 && \textbf{1.04} & 60&60&0&0\\
    &&&  1.46 & \underline{94} & 94 & 0 & 0 && \underline{1.08} & \textbf{90} & \textbf{92} & 0&0\\
    
    \rowcolor{white}& \texttt{camemBERT}&\texttt{BIOMISTRAL} & 1.10 & 27 & 33 & \underline{82}&\underline{88} && 1.40 & 33 & 48 & \underline{81} & \underline{90}\\
    &&& 1.06 & 37& 	\textbf{100} & \textbf{88} & \textbf{90} && 1.56 & 10 & 33 & \textbf{90} & \textbf{92} \\

    \rowcolor{white}& \texttt{DrBERT}&\texttt{BIOMISTRAL} & \underline{1.04} & 14 & 24& 46 & 84 & & 1.20 & 34 & 38 & 74 & 88\\
    \parbox[t]{2mm}{\multirow{-8}{*}{\rotatebox[origin=c]{90}{{\textbf{pRAGe}}}}}&&&  1.08 & 32&\underline{98}& \textbf{88}&\underline{88} & & 1.50 & 14 & 32 & 72 & \underline{88}\\   
    \bottomrule
    \end{tabular}%
    }
    \caption{Manual evaluation comparison of BaseSLMs with pRAGe models for subset of test set. The gray \hlcyan{highlighted rows} correspond to token=50 generation and rest of the rows correspond to token=25.Top scores for each model setups are shown in \textbf{bold} and second highest score is \underline{underlined}.}
    \label{tab:complete-manual}
\end{table*} 

\paragraph{Next, we analyse the results observed from the manual evaluation,} presented in Table \ref{tab:complete-manual}. The human annotation of 50 examples in 24 different configurations (1200 samples)\footnote{See Table 7 for all 24 configurations analyzed.} shows that our fine-tuned version of \texttt{BIOMISTRAL} in inference alone and integrated in the pRAGe pipeline is the best model for \textbf{short answers} (\textbf{90\%} strict correctness). Base \texttt{BIOMISTRAL} is the best model for \textbf{longer answers} in inference alone setting (\textbf{94\%} strict correctness). 
The generated medical paraphrases and explanations should share correct medical knowledge, be informative and concise. This last trait is essential to our study, as we test the generation SLMs for a patient oriented downstream application. In this sense, we further analyzed the generated text according to different criteria:

\paragraph{Correctness of the medical knowledge.} The best model, \texttt{BIOMISTRAL}, non fine-tuned, generated an English word in a French sentence, as seen in example [1]. After finetuning, the model generates a correct answer in French [2]\footnote{The query contained a list of terms (\textit{asthme , mucoviscidose , ventilation mécanique} - asthma , cystic fibrosis , mechanical ventilation), thus explaining the plural form of the generated text.}. In the context of the pRAGe pipeline, \texttt{BIOMISTRAL} gives as well a full French answer [3].
\begin{itemize}
    \item \textbf{non fine-tuned} - \textit{[1] Asthme: maladie où les \textbf{airways} se ferment et se contractent, faisant du bruitlors de l'inspiration et de la respiration} (Asthma: a disease in which the airways close and contract, making noise during inspiration and breathing)
    \item \textbf{fine-tuned} - \textit{[2] maladies respiratoires chroniques et maladies rares respiratoires} (chronic respiratory diseases and rare respiratory diseases)
    \item \textbf{pRAGe (CamemBERT)} - \textit{[3] Asthme: maladie qui fait ressentir des difficultés à respirer, souvent accompagnée de toux et de sifflements.} (Asthma: a disease that makes breathing difficult, often accompanied by coughing and wheezing.)
    
\end{itemize}

\paragraph{While the hallucination \cite{huang2023survey} percentage is low for \texttt{BIOMISTRAL}} alone in zero-shot setting (6\% for fine-tuned and 4\% for vanilla, in the best settings\footnote{The best settings are the \texttt{relaxed correctness} values from \texttt{Token=50}, Table 4.}), \texttt{BARThez} is not adapted for zero-shot inference, as it is a summarization model. In most analyzed cases, it only summarizes the prompt, as the following example shows: \textit{Expliquez-moi le terme médical en mots simples : phobie} (Explain the medical term in simple words: phobia). However, in pRAGe setting, \texttt{BARThez} is capable of summarizing the retrieved documents in a coherent answer. Nevertheless, its hallucination rate is still higher: 50\% for both finetuned and vanilla.  This can be explained by the fact that the model is not adapted for Q\&A tasks, contrary to \texttt{BIOMISTRAL}. In pRAGe, \texttt{BIOMISTRAL} had a lower hallucination rate (10\% in fine-tuned version and 8\% in vanilla version).

However, we observed that \texttt{BIOMISTRAL} alone outperforms its pRAGe counterpart in both fine-tuning and zero-shot setting (+4\% improvement). This results could be explained by the work of \citeauthor{mallen2023not} (\citeyear{mallen2023not}) on the impact of parametric an non-parametric knowledge of GPT language models (LMs) in vanilla and RAG setting depending on fact popularity. The authors' experiments showed that non-parametric memories are more effective for less popular facts (end-tail knowledge) than base LMs. However, they also proved that non-parametric memories can mislead LMs with less factual information. In our experiments, we hypothesize that the popularity of the medical knowledge to be paraphrased and the relevance of the information retrieved from RefoMed-KB can explain the lower performance of \texttt{BIOMISTRAL} in the pRAGe pipeline. Further qualitative analysis is needed to test this hypothesis for medical knowledge.

\paragraph{Short and concise answers.} We analyzed if the generated medical paraphrase or explanation is accurate and concise in two settings: a constraint of 25 and 50 tokens of generated text. In a 50 token setting, the human analysis shows that the best model is \texttt{BIOMISTRAL} accuracy in correct answers in a relaxed and strict setting (96\% ; 94\%), while the second best is its fine-tuned version (94\% ; 90\%). However, the model is not as good in strict correctness in a 25 token setting (68\%). Our fine-tuned version of \texttt{BIOMISTRAL} is better (\textbf{90\%}, up to \textbf{22\%} increase in performance) on strict correctness and conciseness in a 25 tokens setting. 
In the \textbf{pRAGe} pipeline, CamemBERT \texttt{BIOMISTRAL}, both non fine-tuned and fintuned, gave better answers in terms of strict correctness for the 25 token setting (\textbf{82\%; 81\%}). 

\paragraph{Readability for laypeople.}  We analyzed how the \texttt{readability} score was influenced by our different setting experiments. In the short answer setting (25 token), the \texttt{readability} is better with Base \texttt{BIOMISTRAL} (\textbf{1.08}, lower is better). However, even if the readability is good, the answers are incomplete (10\% \texttt{completeness-strict}). Our fine-tuned version of Base \texttt{BIOMISTRAL} improves the completeness of the answer (16\% \texttt{completeness-strict}), while the best pRAGe model, CamemBERT \texttt{BIOMISTRAL} fine-tuned, increases it even further up to \textbf{33\%}. Thus, we see a \textbf{+23\%} improvement in performance (in \texttt{completeness-strict}) with our fine-tuned model in pRAGe.

One aspect that explains the decrease in \texttt{readability} in our fine-tuned models is the higher use of medical terms in the generated answers, as the fine-tuning step with the RefoMed dataset focuses on medical terms. As the goal of this study was to give short and concise paraphrases to a user query, we see that there are advantages of the fine-tuned model: it generates subsentential paraphrases, thus shorter and complete units of meaning (in the 25 tokens setting). Moreover, the fine-tuned model also generates simplified text generations, as observed in the following example where the medical term "osteophyte" is explained by using a subsentential paraphrase in very simple language: "deposits of bone tissue that form on the edges of bones" (original in French \textit{ostéophyte - des dépôts de tissu osseux qui se forment sur les bords des os}).

\begin{table}[ht]

    \centering
    \resizebox{\columnwidth}{!}{
    \begin{tabular}{|r|cc|}
    \toprule
         &  \textbf{Krippendorff’s alpha(nominal)}&	\textbf{\% agreement} \\
         \midrule
         \rowcolor{gray!25}
         \multicolumn{3}{|c|}{token=25} \\
         \multicolumn{3}{|c|}{\textbf{w/o FINE TUNING}} \\
         \midrule
         \textit{completeness}-STRICT&	0.879&	98\% \\
\textit{completeness}-RELAX&	0.649&	90\% \\
\textit{readability}	&1.555	&78\%\\
\midrule
		\multicolumn{3}{|c|}{\textbf{w/ FINE TUNING}} \\
    \midrule
\textit{completeness}-STRICT&	-0.076&	84\%\\
\textit{completeness}-RELAX&	-0.1&	80\%\\
\textit{readability}	&0.132	&68\%\\
\midrule
		\rowcolor{gray!25}
         \multicolumn{3}{|c|}{token=50} \\
         
         \multicolumn{3}{|c|}{\textbf{w/o FINE TUNING}} \\
         \midrule
\textit{completeness}-STRICT&	0.66&	98\%\\
\textit{completeness}-RELAX&	0.003&	64\%\\
\textit{readability}	&0.105&	70\%\\
\midrule
		\multicolumn{3}{|c|}{\textbf{w/ FINE TUNING}} \\
  \midrule
\textit{completeness}-STRICT&	0.105	&70\%\\
\textit{completeness}-RELAX&	1.151&	64\%\\
\textit{readability}&	-0.529&	2\%\\
         \bottomrule
    \end{tabular}}
    \caption{Inter-annotator agreement analysis}
    \label{tab:iaa}

\end{table}

\paragraph{Inter-annotator agreement score.} We computed a Krippendorff's alpha score \cite{krippendorff2018content} for two criteria of the human evaluation: \texttt{completeness} and \texttt{readability}. The annotation was conducted by 3 French linguists annotators: 1 linguist completed a full annotation and 2 other linguists contributed to the second annotation (one annotated Token=25 and the other Token=50 length paraphrases). We show the Krippendorff's alpha score and the percentage agreement in Table 5. The inter-annotator agreement is highest for \texttt{completeness-strict}, for both lengths (98\% agreement), showing syntactic analysis is an easy task for the annotators. However, regarding \texttt{readability}, it is more difficult for the two annotators to agree (78\% to 68\% agreement), meaning that the medical knowledge of the annotators can influence the readability level of annotations.

\begin{figure}
    \centering
    \includegraphics[scale=0.5, trim=0cm 0cm 0cm 0cm,width=\columnwidth]{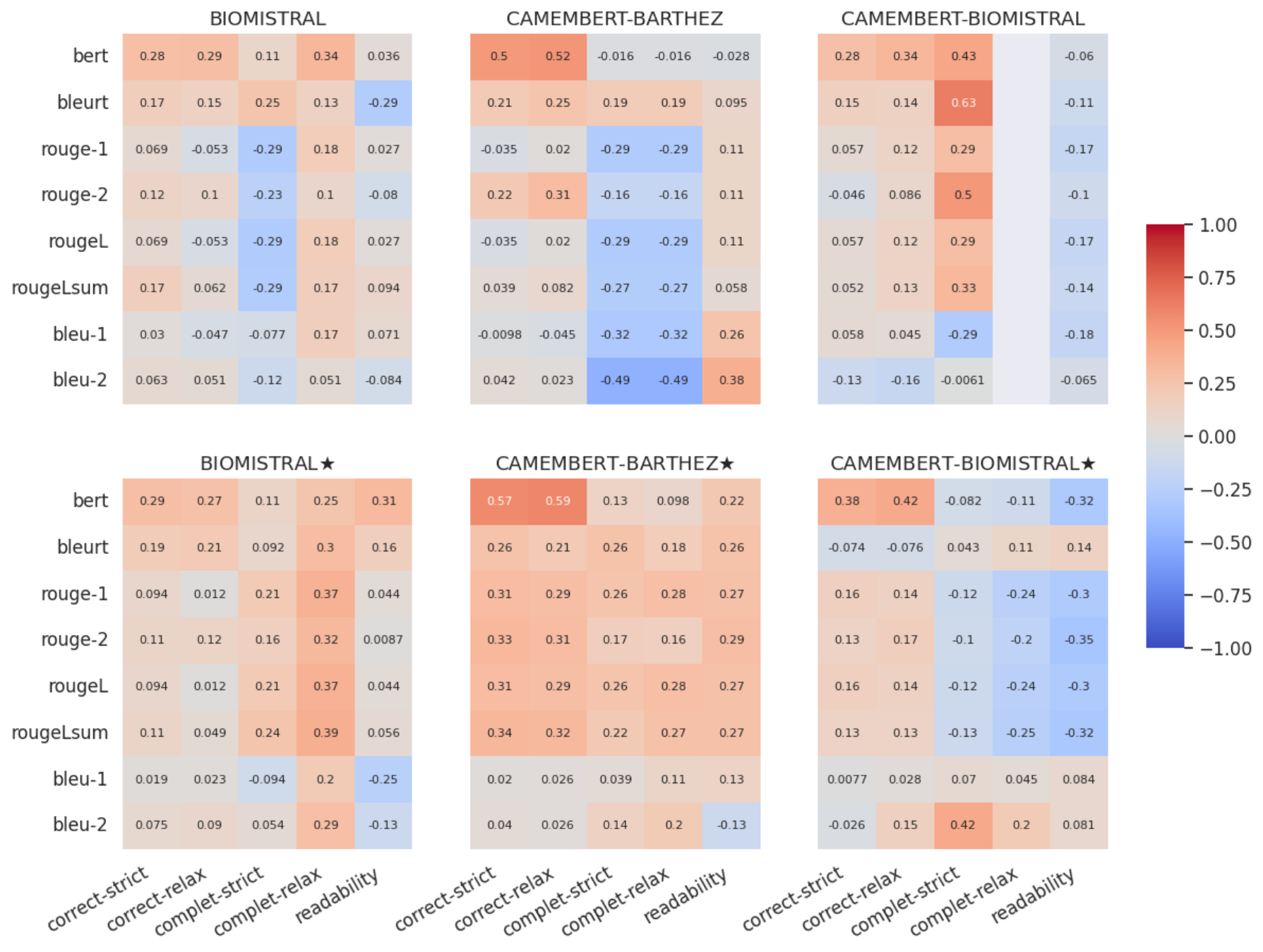}
    \caption{Correlation Heatmap between Automatic evaluation metrics (y-axis) and Manual evaluation metrics (x-axis). The $\bigstar$ symbol denotes  configurations with finetuned SLM.}
    \label{fig:enter-label}
\end{figure}
\paragraph{Automatic evaluation VS Manual evaluation.} In Figure 3 (which merges values from Table 3 and Table 4), we show the correlation heatmap between automatic metric scores and manual metrics scores for \textit{best} SLM and pRAGe configurations. The top row corresponds to configuration without finetuning and the bottom row corresponds to configurations with finetuned SLM backbone. \texttt{rouge} metrics positively corrected with \texttt{correctness} for the generation with finetuning which is intuitive as finetuning allows the pRAGe setups to generate in a more similar style as the gold reference (RefoMed). \texttt{bert} and \texttt{bleurt} scores remain unaffected with finetuning as the semantic similarity of generation can saturate unless lexical similarity increases. Finally, we notice the \texttt{readability} aspect of SLMs is differently affected from pRAGe setup as it is not as trivial as lexical closeness. 

\section{Conclusion and Future Work}

We presented pRAGe, a pipeline for retrieval, generation and evaluation of medical paraphrases in a patient/lay person oriented downstream application. We showed that finetuning \texttt{BIOMISTRAL} with the RefoMed dataset increases the ability of the model to generate short, concise and correct subsentential paraphrases. The pRAGe pipeline helps increase the scientific grounding of text generation via LLMs for medical domain. Our work intends to help bridge the gap between scientific medical knowledge and lay people. 

Further work will include testing another French language models, such as Vigogne or Claire in similar settings, finetuned and in pRAGe. We are also planning to conduct an extensive annotation campaign with specialists of the medical domain to asses a fair inter-agreement score on \texttt{correctness} and further explore how the generation is affected by the choice of knowledge base used.  


\section*{Ethics Statement}

We chose to work only with open source LLMs and RAG models in order to allow reproducibility and replicability of our scientific method and of our results. We consider it is very important to democratize access to the adaptation of LLMs to specific downstream tasks by improving results with smaller language models (SLMs) that can run of less GPUs. We share our experimental setup with the scientific community.

\section*{Acknowledgments}

We thank the anonymous reviewers for their useful comments
which allowed to improve this paper. And, we are also thankful to Timothee Mickus and Fanny Ducel for helping with the annotations. The experiments presented in this paper were carried out using the Grid’5000 testbed, supported by a scientific interest group hosted by Inria and including CNRS, RENATER and several Universities as well as other organizations (see \texttt{https://www.grid5000.fr}). This work has been partially funded by the IDEX/I-SITE initiative ``Lorraine Université d’Excellence (LUE)'' through its call for interdisciplinary projects.

\bibliography{custom}


\appendix
\onecolumn
\section{Appendix}
\label{sec:appendix}

\begin{table}[]
    \centering
    \begin{tabular}{|p{14cm}|}
    \toprule
         Vous êtes un expert en médecine. Utilisez les informations 
suivantes pour répondre à la question de l'utilisateur par une paraphrase, 
une explication ou une courte définition.

Si vous ne connaissez pas la réponse, dites simplement que vous ne savez pas,
n'essayez pas d'inventer une réponse.

\textbf{Contexte}: {context}

\textbf{Question}: {question}

Ne renvoyez que la réponse utile. La réponse doit être claire, concise et facile
à comprendre pour le grand public.

\textbf{Réponse utile} :  \\
\bottomrule
    \end{tabular}
    \caption{Initial prompt for base SLM experiments.}
    \label{tab:firstprompt}
\end{table}

\subsection{Sample prompt for pRAGe} We use the following prompt (See Table \ref{tab:firstprompt}) to start our experiment and then tune it.
\subsection{Complete Automatic Evaluation} We provide in Table \ref{tab:final-test-reports} detail results for all the configuration we consider during our experiments.

\subsection{Manual Evaluation Examples} We provide in Table  \ref{tab:biom-manual}, \ref{tab:camem-manual} samples from BIOMISTRAL and BARTHEZ SLMs and pRAGe configurations and their manual evaluation annotation which are used to create the final table (See Table \ref{tab:complete-manual}).


%

\begin{table*}[]
    \centering
    \resizebox{\textwidth}{!}{%
    \begin{tabular}{p{5cm} | c |c c ccccccccc}
    \toprule
        Setup & TOKEN& -bleu&-bert&-bleurt&-rouge1&-rouge2&-rougL&-rougeLsum&-bleu-p1&-bleu-p2&-bleu-p3&-bleu-p4\\
    \midrule
    \rowcolor{gray!25}
    && \multicolumn{11}{c}{w/o FINE TUNING}\\
    {\textbf{BARTHEZ}} & 25 &  $0.00_{{0.00}}$&$0.63_{{0.03}}$&$0.10_{{0.10}}$&$0.07_{{0.08}}$&$0.01_{{0.03}}$&$0.07_{{0.08}}$&$0.06_{{0.07}}$&$0.04_{{0.06}}$&$0.00_{{0.01}}$&$0.00_{{0.00}}$&$0.00_{{0.00}}$\\
    & 50 &$0.00_{{0.00}}$&$0.63_{{0.03}}$&$0.10_{{0.10}}$&$0.07_{{0.08}}$&$0.01_{{0.03}}$&$0.07_{{0.08}}$&$0.06_{{0.07}}$&$0.04_{{0.06}}$&$0.00_{{0.01}}$&$0.00_{{0.00}}$&$0.00_{{0.00}}$\\
    \midrule
    {\textbf{BIOMISTRAL}} & 25 & $0.00_{{0.02}}$&$0.70_{{0.06}}$&$0.15_{{0.15}}$&$0.20_{{0.16}}$&$0.07_{{0.12}}$&$0.20_{{0.16}}$&$0.17_{{0.14}}$&$0.11_{{0.12}}$&$0.03_{{0.08}}$&$0.01_{{0.06}}$&$0.01_{{0.05}}$\\
    & 50 &$0.00_{{0.03}}$&$0.68_{{0.06}}$&$0.16_{{0.15}}$&$0.18_{{0.13}}$&$0.06_{{0.09}}$&$0.18_{{0.13}}$&$0.14_{{0.11}}$&$0.08_{{0.08}}$&$0.02_{{0.05}}$&$0.01_{{0.03}}$&$0.00_{{0.03}}$\\
    \midrule
    {\textbf{CAMEMBERT+BARTHEZ}} & 25 & $0.00_{{0.00}}$&$0.65_{{0.05}}$&$0.07_{{0.09}}$&$0.12_{{0.10}}$&$0.02_{{0.05}}$&$0.12_{{0.10}}$&$0.10_{{0.08}}$&$0.05_{{0.07}}$&$0.00_{{0.02}}$&$0.00_{{0.01}}$&$0.00_{{0.00}}$ \\
    & 50 &$0.00_{{0.00}}$&$0.65_{{0.05}}$&$0.11_{{0.11}}$&$0.12_{{0.10}}$&$0.02_{{0.05}}$&$0.12_{{0.10}}$&$0.10_{{0.08}}$&$0.05_{{0.06}}$&$0.00_{{0.02}}$&$0.00_{{0.01}}$&$0.00_{{0.00}}$ \\
    \midrule
    {\textbf{DRBERT+BARTHEZ}} & 25 &$0.00_{{0.00}}$&$0.64_{{0.03}}$&$0.02_{{0.06}}$&$0.10_{{0.09}}$&$0.00_{{0.02}}$&$0.10_{{0.09}}$&$0.08_{{0.06}}$&$0.04_{{0.06}}$&$0.00_{{0.01}}$&$0.00_{{0.00}}$&$0.00_{{0.00}}$\\
    & 50&$0.00_{{0.00}}$&$0.65_{{0.04}}$&$0.05_{{0.07}}$&$0.11_{{0.09}}$&$0.01_{{0.02}}$&$0.11_{{0.09}}$&$0.09_{{0.07}}$&$0.05_{{0.06}}$&$0.00_{{0.01}}$&$0.00_{{0.00}}$&$0.00_{{0.00}}$\\
    \midrule
    {\textbf{CAMEMBERT+BIOMISTRAL}} & 25 &$0.01_{{0.06}}$&$0.69_{{0.06}}$&$0.14_{{0.15}}$&$0.19_{{0.17}}$&$0.08_{{0.14}}$&$0.19_{{0.17}}$&$0.17_{{0.15}}$&$0.12_{{0.14}}$&$0.04_{{0.12}}$&$0.02_{{0.11}}$&$0.02_{{0.10}}$\\
    & 50 & $0.00_{{0.03}}$&$0.68_{{0.06}}$&$0.17_{{0.15}}$&$0.18_{{0.14}}$&$0.06_{{0.10}}$&$0.18_{{0.14}}$&$0.15_{{0.12}}$&$0.08_{{0.09}}$&$0.02_{{0.05}}$&$0.01_{{0.04}}$&$0.01_{{0.04}}$ \\
    \midrule
   {\textbf{ DRBERT+BIOMISTRAL}} & 25 &$0.00_{{0.02}}$&$0.69_{{0.06}}$&$0.14_{{0.15}}$&$0.18_{{0.17}}$&$0.07_{{0.13}}$&$0.18_{{0.17}}$&$0.16_{{0.16}}$&$0.11_{{0.12}}$&$0.03_{{0.08}}$&$0.02_{{0.06}}$&$0.01_{{0.05}}$\\
    & 50 & $0.00_{{0.02}}$&$0.68_{{0.06}}$&$0.17_{{0.15}}$&$0.17_{{0.13}}$&$0.05_{{0.09}}$&$0.17_{{0.13}}$&$0.14_{{0.12}}$&$0.08_{{0.08}}$&$0.02_{{0.05}}$&$0.01_{{0.04}}$&$0.00_{{0.03}}$\\
    \midrule
    \rowcolor{gray!25}
    && \multicolumn{11}{c}{w/ FINE TUNING}\\
    {\textbf{BARTHEZ}$\bigstar$} & 25 & $0.00_{{0.00}}$&$0.62_{{0.02}}$&$0.05_{{0.08}}$&$0.11_{{0.08}}$&$0.01_{{0.02}}$&$0.11_{{0.08}}$&$0.09_{{0.06}}$&$0.06_{{0.07}}$&$0.00_{{0.01}}$&$0.00_{{0.00}}$&$0.00_{{0.00}}$ \\
    & 50 &$0.00_{{0.01}}$&$0.63_{{0.03}}$&$0.09_{{0.09}}$&$0.12_{{0.08}}$&$0.01_{{0.04}}$&$0.12_{{0.08}}$&$0.09_{{0.06}}$&$0.07_{{0.07}}$&$0.01_{{0.02}}$&$0.00_{{0.02}}$&$0.00_{{0.01}}$\\
    \midrule
    {\textbf{BIOMISTRAL}$\bigstar$} & 25 & $0.00_{{0.00}}$&$0.72_{{0.07}}$&$0.15_{{0.17}}$&$0.22_{{0.17}}$&$0.09_{{0.13}}$&$0.22_{{0.17}}$&$0.20_{{0.16}}$&$0.14_{{0.13}}$&$0.04_{{0.07}}$&$0.01_{{0.04}}$&$0.00_{{0.02}}$\\
    & 50 & $0.00_{{0.00}}$&$0.69_{{0.07}}$&$0.16_{{0.16}}$&$0.18_{{0.13}}$&$0.07_{{0.10}}$&$0.18_{{0.13}}$&$0.16_{{0.12}}$&$0.10_{{0.10}}$&$0.02_{{0.04}}$&$0.01_{{0.02}}$&$0.00_{{0.01}}$\\
    \midrule
    {\textbf{CAMEMBERT+BARTHEZ}$\bigstar$} & 25 &$0.00_{{0.00}}$&$0.65_{{0.05}}$&$0.05_{{0.09}}$&$0.12_{{0.10}}$&$0.02_{{0.05}}$&$0.12_{{0.10}}$&$0.10_{{0.08}}$&$0.06_{{0.07}}$&$0.01_{{0.02}}$&$0.00_{{0.01}}$&$0.00_{{0.00}}$\\
    & 50  &$0.00_{{0.01}}$&$0.64_{{0.05}}$&$0.10_{{0.10}}$&$0.12_{{0.10}}$&$0.02_{{0.05}}$&$0.12_{{0.10}}$&$0.10_{{0.08}}$&$0.06_{{0.07}}$&$0.01_{{0.02}}$&$0.00_{{0.01}}$&$0.00_{{0.01}}$\\
    \midrule
    {\textbf{DRBERT+BARTHEZ}$\bigstar$} & 25 &$0.00_{{0.00}}$&$0.64_{{0.03}}$&$0.01_{{0.04}}$&$0.13_{{0.10}}$&$0.01_{{0.03}}$&$0.13_{{0.10}}$&$0.10_{{0.07}}$&$0.06_{{0.07}}$&$0.00_{{0.01}}$&$0.00_{{0.00}}$&$0.00_{{0.00}}$\\
    & 50&$0.00_{{0.00}}$&$0.64_{{0.04}}$&$0.05_{{0.07}}$&$0.12_{{0.09}}$&$0.01_{{0.03}}$&$0.12_{{0.09}}$&$0.09_{{0.06}}$&$0.05_{{0.06}}$&$0.00_{{0.01}}$&$0.00_{{0.00}}$&$0.00_{{0.00}}$\\
    \midrule
    {\textbf{CAMEMBERT+BIOMISTRAL}$\bigstar$} & 25 &$0.00_{{0.00}}$&$0.60_{{0.04}}$&$0.13_{{0.11}}$&$0.09_{{0.06}}$&$0.02_{{0.03}}$&$0.09_{{0.06}}$&$0.07_{{0.05}}$&$0.03_{{0.03}}$&$0.01_{{0.01}}$&$0.00_{{0.00}}$&$0.00_{{0.00}}$\\
    & 50 & $0.00_{{0.00}}$&$0.60_{{0.05}}$&$0.16_{{0.11}}$&$0.09_{{0.06}}$&$0.02_{{0.03}}$&$0.09_{{0.06}}$&$0.07_{{0.05}}$&$0.03_{{0.03}}$&$0.01_{{0.01}}$&$0.00_{{0.00}}$&$0.00_{{0.00}}$  \\
    \midrule
   {\textbf{ DRBERT+BIOMISTRAL}$\bigstar$} & 25 &$0.00_{{0.00}}$&$0.59_{{0.04}}$&$0.12_{{0.15}}$&$0.08_{{0.06}}$&$0.02_{{0.02}}$&$0.08_{{0.06}}$&$0.07_{{0.04}}$&$0.03_{{0.03}}$&$0.00_{{0.01}}$&$0.00_{{0.00}}$&$0.00_{{0.00}}$\\
    & 50 & $0.00_{{0.00}}$&$0.60_{{0.04}}$&$0.14_{{0.15}}$&$0.08_{{0.06}}$&$0.02_{{0.02}}$&$0.08_{{0.06}}$&$0.07_{{0.04}}$&$0.03_{{0.02}}$&$0.00_{{0.01}}$&$0.00_{{0.00}}$&$0.00_{{0.00}}$\\
    \bottomrule
           \end{tabular}}
    \caption{Detailed Experiment report on \underline{test set} following equation \ref{eqn:rag-eval} for each metric. The $\bigstar$ symbol denotes  configurations with finetuned SLM.}
    \label{tab:final-test-reports}
\end{table*}

\begin{table*}[ht]
    \centering
    \resizebox{0.9\textwidth}{!}{%
    \begin{tabular}{cc c p{6cm} p{6cm}}
    \toprule
         && & \multicolumn{2}{c}{Base SLM : BIOMISTRAL}\\
         \cline{4-5}
         &  TOKEN& &\small{NON-FINETUNED} & \small{FINETUNED} \\
         
         \midrule
          &&  \multicolumn{3}{g}{PROMPT + \textbf{ostéophyte}} \\
          
         SLM & 25& \texttt{fr} &\small{ Un ostéophyte est une formation osseuse qui pousse à partir d'une articulation ou d} & \small{des dépôts de tissu osseux qui se forment sur les bords des os[1} \\
         && \texttt{en} &  (\small{An osteophyte is a bony formation that grows from a joint or }) & (\small{deposits of bone tissue that form on the edges of bones[1})\\
         
        &&\texttt{[TAG]} &\small{\texttt{Cr-R:1; Cr-S:1; Cm-R:0; Cm-S:0; Rd:2.}} &\small{\texttt{Cr-R:1; Cr-S:1; Cm-R:1. Cm-S:0; Rd:1.}} \\       
         \cline{4-5}
         
          &&   \multicolumn{3}{g}{PROMPT + \textbf{asthme, mucoviscidose, ventilation mécanique}} \\
          &&\texttt{fr} &\small{Asthme: maladie où les airways se ferment et se contractent, faisant du bruit lors} & \small{maladies respiratoires chroniques et maladies rares respiratoires (M)} \\
         && \texttt{en} &  (\small{Asthma: a disease in which the airways close and contract, making noise as}) & (\small{chronic respiratory diseases and rare respiratory diseases (M)} \\
         &&\texttt{[TAG]} &\small{\texttt{Cr-R:1; Cr-S:0; Cm-R:0; Cm-S:0; Rd:1.}} &\small{\texttt{Cr-R:1; Cr-S:1; Cm-R:1; Cm-S:0; Rd:1.}}  \\
         \cline{2-5}
          &&  \multicolumn{3}{g}{PROMPT + \textbf{ostéophyte}} \\
            & 50 & \texttt{fr} &\small{Un ostéophyte est une formation osseuse qui pousse à partir d'une articulation ou d'une jointure. Il peut être formé en réponse à une blessure ou à une inflammation, mais} & \small{des dépôts de tissu osseux qui se forment sur les bords des os [1] [2] [3] [4] [5] [6] [7]} \\
         && \texttt{en} & (\small{An osteophyte is a bone formation that grows from a joint or articulation. It may be formed in response to injury or inflammation, but}) & (\small{deposits of bone tissue that form on the edges of bones [1] [2] [3] [4] [5] [6] [7]}) \\
         &&\texttt{[TAG]} &\small{\texttt{Cr-R:1; Cr-S:1; Cm-R:1; Cm-S:0; Rd:2.}} &\small{\texttt{Cr-R:1; Cr-S:1; Cm-R:1; Cm-S:0; Rd:1.}} \\
         \cline{4-5}
         
          &&   \multicolumn{3}{g}{PROMPT + \textbf{asthme, mucoviscidose, ventilation mécanique}} \\
         && \texttt{fr} &\small{Asthme: maladie où les airways se ferment et se contractent, faisant du bruit lors de l'inspiration et de la respiration. Mucoviscidose: maladie où} & \small{maladies respiratoires chroniques et maladies rares respiratoires (MRR) (MRR) (MRR) (MRR) (MRR) (M} \\
         && \texttt{en} & (\small{Asthma: a disease in which the airways close and contract, making noise when breathing in and out. Cystic fibrosis: a disease in which}) & (\small{chronic respiratory diseases and rare respiratory diseases (MRR) (MRR) (MRR) (MRR) (MRR) (MRR) (M)} \\
         &&\texttt{[TAG]} &\small{\texttt{Cr-R:1; Cr-S:0; Cm-R:1; Cm-S:0; Rd:1.}} &\small{\texttt{Cr-R:1; Cr-S:1; Cm-R:1; Cm-S:0; Rd:1.}} \\
         \midrule
             &&&  \multicolumn{2}{c}{pRAGe : BIOMISTRAL}\\
         \cline{4-5}
          &&  \multicolumn{3}{g}{PROMPT + \textbf{ostéophyte}} \\
    C.BERT & 25 & \texttt{fr}&\small{Un ostéophyte est une formation osseuse qui pousse autour d'une articulation. Il} & \small{est une formation osseuse qui se développe sur les os entourant une articulation.} \\
         && \texttt{en} &  (\small{An osteophyte is a bone formation that grows around a joint. It}) & (\small{is a bone formation that develops on the bones surrounding a joint.)} \\
         &&\texttt{[TAG]} &\small{\texttt{Cr-R:1; Cr-S:1; Cm-R:1; Cm-S:0; Rd:2.}} &\small{\texttt{Cr-R:1; Cr-S:1; Cm-R:1; Cm-S:1; Rd:1.}} \\    
         \cline{4-5}
          &&  \multicolumn{3}{g}{PROMPT + \textbf{asthme, mucoviscidose, ventilation mécanique}} \\
         && \texttt{fr} &\small{Asthme: maladie qui fait ressentir des difficultés à respirer, souvent accompagnée de} & \small{est une maladie génétique qui affecte les glandes salivaires et les glandes lacry} \\
         && \texttt{en} &  (\small{Asthma: a disease that makes it difficult to breathe, often accompanied b}) & (\small{is a genetic disease that affects the salivary and lacrimal glan)} \\
         &&\texttt{[TAG]} &\small{\texttt{Cr-R:1; Cr-S:1; Cm-R:0; Cm-S:0; Rd:1.}} &\small{\texttt{Cr-R:1; Cr-S:0; Cm-R:0; Cm-S:0; Rd:1.}} \\
    \cline{2-5}
          &&  \multicolumn{3}{g}{PROMPT + \textbf{ostéophyte}} \\
    & 50 & \texttt{fr} &\small{Un ostéophyte est une formation osseuse qui pousse autour d'une articulation. Il peut être causé par une maladie appelée arthrose, qui fait que de la matière osseuse} & \small{est une formation osseuse qui se développe sur les os entourant une articulation. Les ostéophytes peuvent être symptomatiques ou asymptomatiques. Lors} \\
         && \texttt{en} & (\small{An osteophyte is a bone formation that grows around a joint. It can be caused by a disease called arthritis, which causes bone material to)} & (\small{is a bone formation that develops on the bones surrounding a joint. Osteophytes may be symptomatic or asymptomatic. At)} \\
         
         &&\texttt{[TAG]} &\small{\texttt{Cr-R:1; Cr-S:1; Cm-R:1; Cm-S:0; Rd:2.}} &\small{\texttt{Cr-R:1; Cr-S:1; Cm-R:1; Cm-S:0; Rd:1.}} \\
         \cline{4-5}
          &&  \multicolumn{3}{g}{PROMPT + \textbf{asthme, mucoviscidose, ventilation mécanique}} \\
         && \texttt{fr} &\small{Asthme: maladie qui fait ressentir des difficultés à respirer, souvent accompagnée de toux et de sifflements. Mucoviscidose: maladie génétique qui affect} & \small{est une maladie génétique qui affecte les glandes salivaires et les glandes lacrymales, provoquant une production excessive de mucus. Cette maladie peut également affecter les voies} \\
         
         && \texttt{en} &  (\small{Asthma: a disease that makes breathing difficult, often accompanied by coughing and wheezing. Cystic fibrosis: genetic disease that affects}) & (\small{is a genetic disorder that affects the salivary and lacrimal glands, causing excessive mucus production. This disease can also affect the [tracks])} \\
         
         &&\texttt{[TAG]} &\small{\texttt{Cr-R:1; Cr-S:1; Cm-R:1; Cm-S:0; Rd:1.}} &\small{\texttt{Cr-R:1; Cr-S:1; Cm-R:1; Cm-S:0; Rd:1.}} \\
         \bottomrule
    \end{tabular}
    }
    \caption{Examples of generated answers from the different BioMistral configurations. The human annotation tags stand for our evaluation metrics: \texttt{correctness-relaxed (Cr-R)}, \texttt{correctness-strict (Cr-S)}, \texttt{completeness-relaxed (Cm-R)}, \texttt{completeness-strict (Cm-S)}, and \texttt{readability (Rd)}, and \textit{C.BERT} for CamemBERT sentence-embedding. PROMPT refers to the prompt syntax in Fig.\ref{fig:prompt}}
    \label{tab:biom-manual}
\end{table*}
\begin{table*}[]
    \centering
    \resizebox{\textwidth}{!}{%
    \begin{tabular}{cc c p{6cm} p{6cm}}
    \toprule
         && & \multicolumn{2}{c}{pRAGe : BARTHEZ}\\
         \cline{4-5}
         \rowcolor{white}& TOKEN & &\small{NON-FINETUNED} & \small{FINETUNED} \\
        
         \midrule
         \cline{4-5}
         &&  \multicolumn{3}{g}{PROMPT + \textbf{ostéophyte}} \\
    \rowcolor{white}C.BERT & 25 & \texttt{fr}&\small{L'arthrose est une affection dégénérative des articulations qui fait que de la matière osseuse est anormalement produite} & \small{L'ostéophytose désigne le phénomène d'apparition d'un ostéophyte sur une articulation, lors} \\
         \rowcolor{white}&& \texttt{en} &  (\small{Osteoarthritis is a degenerative joint disorder in which bone material is abnormally produced.}) & (\small{Osteophytosis refers to the phenomenon of the appearance of an osteophyte on a joint, when)} \\
         \rowcolor{white}&&\texttt{[TAG]} &\small{\texttt{Cr-R:0; Cr-S:0; Cm-R:1; Cm-S:0; Rd:1.}} &\small{\texttt{Cr-R:1; Cr-S:1; Cm-R:0; Cm-S:0; Rd:2.}} \\    
         \cline{4-5}
          &&  \multicolumn{3}{g}{PROMPT + \textbf{asthme, mucoviscidose, ventilation mécanique}} \\
         && \texttt{fr} &\small{Ces descriptions se focalisent le plus souvent sur les problèmes digestifs, stéatorrhée et troubles pancréatiques,} & \small{Notes et références Voir aussi Articles connexes BradypnéeDyspnéePolypnéeT} \\
         && \texttt{en} &  (\small{These descriptions most often focus on digestive problems, steatorrhea and pancreatic disorders,}) & (\small{Notes and references See also Related articles BradypneaDyspneaPolypneaT)} \\
          &&\texttt{[TAG]} &\small{\texttt{Cr-R:0; Cr-S:0; Cm-R:0; Cm-S:0; Rd:2.}} &\small{\texttt{Cr-R:0; Cr-S:0; Cm-R:0; Cm-S:0; Rd:3.}} \\
    \cline{2-5}
          &&  \multicolumn{3}{g}{PROMPT + \textbf{ostéophyte}} \\
    & 50 & \texttt{fr} &\small{L'arthrose est une affection dégénérative des articulations qui fait que de la matière osseuse est anormalement produite à l'extérieur de l'os.} & \small{L'ostéophytose est une affection dégénérative des articulations qui fait que de la matière osseuse est anormalement produite à l'extérieur de l'os, au détriment de la membrane blanchâtre fibreuse qui entoure l'os} \\
         && \texttt{en} & (\small{Osteoarthritis is a degenerative joint disease in which abnormal bone material is produced outside the bone.)} & (\small{Osteophytosis is a degenerative joint disorder in which bone material is abnormally produced on the outside of the bone, to the detriment of the whitish fibrous membrane that surrounds the bone.} \\
         &&\texttt{[TAG]} &\small{\texttt{Cr-R:1; Cr-S:0; Cm-R:1; Cm-S:1; Rd:1.}} &\small{\texttt{Cr-R:1; Cr-S:1; Cm-R:1; Cm-S:1; Rd:2.}} \\
         &&  \multicolumn{3}{g}{PROMPT + \textbf{asthme, mucoviscidose, ventilation mécanique}} \\
         && \texttt{fr} &\small{Ces descriptions se focalisent le plus souvent sur les problèmes digestifs, stéatorrhée et troubles pancréatiques.} & \small{Notes et références Voir aussi Articles connexes BradypnéeDyspnéePolypnéeTachypnéeSyndrome d'apnées du sommeilApnée (sport)} \\
         && \texttt{en} &  (\small{These descriptions most often focus on digestive problems, steatorrhea and pancreatic disorders.}) & (\small{Notes and references See also Related articles BradypneaDyspneaPolypneaTachypneaSleep apnea syndromeApnea (sport)))} \\
         &&\texttt{[TAG]} &\small{\texttt{Cr-R:0; Cr-S:0; Cm-R:1; Cm-S:1; Rd:2.}} &\small{\texttt{Cr-R:0; Cr-S:0; Cm-R:0; Cm-S:0; Rd:3.}} \\     
         \bottomrule
    \end{tabular}
    }
    \caption{Examples of generated answers from BARTHEZ in the CamemBERT pRAGe setup. The human annotation tags stand for our evaluation metrics: \texttt{correctness-relaxed (Cr-R)}, \texttt{correctness-strict (Cr-S)}, \texttt{completeness-relaxed (Cm-R)}, \texttt{completeness-strict (Cm-S)}, and \texttt{readability (Rd)}, and \textit{C.BERT} for CamemBERT sentence-embedding.PROMPT refers to the prompt syntax in Fig.\ref{fig:prompt}}
    \label{tab:camem-manual}
\end{table*}

\end{document}